\documentclass[10pt,twocolumn,letterpaper]{article}

\usepackage{wacv}              

\usepackage{graphicx}
\usepackage{amsmath}
\usepackage{amssymb}
\usepackage{booktabs}

\usepackage{times}
\usepackage{epsfig}
\usepackage{multirow}
\usepackage[table]{xcolor}
\usepackage{gensymb}
\usepackage{balance}
\usepackage[accsupp]{axessibility}  

\usepackage[pagebackref,breaklinks,colorlinks]{hyperref}

\usepackage[capitalize]{cleveref}
\crefname{section}{Sec.}{Secs.}
\Crefname{section}{Section}{Sections}
\Crefname{table}{Table}{Tables}
\crefname{table}{Tab.}{Tabs.}

\def\wacvPaperID{12} 
\def\confName{WACV}
\def\confYear{2024}

\begin{document}

\title{Iris Presentation Attack: Assessing the Impact of Combining Vanadium Dioxide Films with Artificial Eyes}

\author{Darshika Jauhari, Renu Sharma, Cunjian Chen, Nelson Sepulveda, Arun Ross \\
Michigan State University\\
{\tt\small \{jauharid,sharma90,cunjian,sepulve6,rossarun\}@msu.edu}
}
\maketitle

\begin{abstract}
Iris recognition systems, operating in the near infrared spectrum (NIR), have demonstrated vulnerability to presentation attacks, where an adversary uses artifacts such as cosmetic contact lenses, artificial eyes or printed iris images in order to circumvent the system. At the same time, a number of effective presentation attack detection (PAD) methods have been developed. These methods have demonstrated success in detecting artificial eyes (e.g., fake Van Dyke eyes) as presentation attacks. In this work, we seek to alter the optical characteristics of artificial eyes by affixing Vanadium Dioxide ($\mathit{VO_2}$) films on their surface in various spatial configurations. $\mathit{VO_2}$ films can be used to selectively transmit NIR light and can, therefore, be used to regulate the amount of NIR light from the object that is captured by the iris sensor. We study the impact of such images produced by the sensor on two state-of-the-art iris PA detection methods. We observe that the addition of $\mathit{VO_2}$ films on the surface of artificial eyes can cause the PA detection methods to misclassify them as bonafide eyes in some cases. This represents a vulnerability that must be systematically analyzed and effectively addressed.
\end{abstract}

\section{Introduction}
\begin{figure}[h]
  \includegraphics[width=\linewidth]{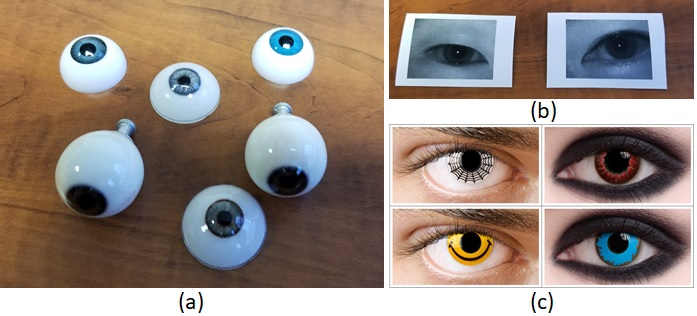}
  \caption{Examples of PAI used to launch presentation attacks (PAs) on the iris modality: (a) plastic eyes, (b) printed images, and (c) cosmetic contacts \cite{ContactsPicRef}.}
  \label{fig:SamplePAs}
\end{figure}

Iris recognition systems use the texture of the iris in order to recognize individuals \cite{Daugman1999}. A typical iris recognition system operates in the near-infrared (NIR) spectrum. There are several reasons for using NIR sensors to acquire an image of the iris: (a) NIR illumination is non-invasive and, unlike visible spectrum lighting, does not excite the pupil; and (b) NIR illumination can be used to elicit the texture of even dark-colored irides since it can penetrate the multi-layered iris more effectively than visible spectrum lighting. Despite their success in a number of real-world applications, iris systems are vulnerable to number of attacks \cite{Nalini2001}, including presentation attacks (PAs)~\cite{Czajka2018, Boyd2022}. A presentation attack occurs when an adversary presents a fake or altered trait to the sensor in order to obfuscate their own identity, spoof another person's identity or to create a virtual identity. The biometric characteristics or materials used to launch a presentation attack are referred to as Presentation Attack Instruments (PAI). Examples of PAIs in the case of the iris modality include printed iris images \cite{ Daugman1999, Czajka2013, Raghavendra2015, Hoffman2018}, plastic, glass, or doll eyes \cite{Hoffman2018, Lee2006}, cosmetic contact lenses \cite{Raghavendra2017, Yadav2014, Hughes2013, ChenR18}, a video display of an eye image \cite{Raja2015, Czajka2018, LivDet2020}, cadaver eyes \cite{Marcel2019, Czajka2018, LivDet2020}, robotic eye models \cite{Komogortsev2015} holographic eye images \cite{Pacut2006} and synthesized irises \cite{LivDet2023}. A few examples of iris PAIs are shown in Fig. \ref{fig:SamplePAs}.  


Among all the attacks described above, iris pattern printed on a paper is perhaps one of the easiest ones. The efficacy of this type of attack depends on a number of factors including the choice of printer (inkjet or laserjet), paper (matte, glossy, photographic, butter, white, recycled or cardboard), resolution (600 or 1,200 dpi), image type (grayscale or color), configuration (with or without pupil cutout), and sensing device (IrisPass, IrisAccess, or IrisGuard). In the LivDet-Iris 2013  competition \cite{LivDet2013}, a combination of two different printers, two commercial iris sensors and matte paper were used. Later, the dataset was extended in the LivDet-Iris 2015 \cite{LivDet2015}, LivDet-Iris 2017 \cite{LivDet2017}, LivDet-Iris 2020 \cite{LivDet2020} and LivDet-Iris 2023 \cite{LivDet2023} competitions by including more variations in the resolution, contrast and texture of the printed irides. In \cite{LivDet2020}, various add-ons were applied to printed paper, including transparent domes and textured as well as clear contact lenses.

The use of cosmetic contact lens as a PAI poses an even greater challenge than the prints, since the former has significantly more manufacturers, brands, and colors \cite{LivDet2017, LivDet2020}. In \cite{LivDet2013}, 22 types of patterned contact lenses were collected, which was later increased to 57 types with different texture patterns \cite{Sun2014}. In \cite{LivDet2015}, 20 different varieties of cosmetic contacts were used to generate iris PA samples. This was later extended by adding samples from the Notre Dame subset, which contained five different brands of textured contact lenses, and the IITD-WVU subset, which contained four manufacturers and six colors \cite{LivDet2017}. In \cite{LivDet2020}, three different brands of cosmetic contacts (Johnson \& Johnson, Ciba Vision, and Bausch \& Lomb) were captured using the LG IrisAccess 4000 and IrisGuard AD100 under various illumination setups (two different illuminants in LG4000 and six different illuminants in AD100).

In addition to the printed and cosmetic contact attacks, the replay or display attack also poses a challenge. In this type of attack, a previously captured iris image or video is presented to an iris sensor via a display media. However, most modern computers, laptops and mobile phone screens do not necessarily emit NIR light. So this type of attack has been predominantly tested on iris systems operating in the visible spectrum \cite{Raja2015, Czajka2018}. However, the display of certain Kindle devices emit NIR light and, consequently, can be more easily imaged using NIR iris sensors \cite{Hoffman2019, LivDet2020}. 

A plastic or prosthetic eye is a highly viable PAI, but has not been as extensively explored in the literature, unlike some of the other PAs. Variants of such artificial eyes can be designed using different materials like Poly Methyl Meta Acrylate \cite{Lee2006}, glass or plastic. Lee et al. \cite{Lee2006} created three different-colored artificial eyes (blue, gray and dark brown). Sun et al. \cite{Sun2014} selected 40 different subjects iris images from the UPOL database \cite{UPOL} and printed them on plastic eyeball models. Hoffman et al. \cite{Hoffman2019} collected images of fake eyes using three different plastic/glass eye brands and 10 distinct colors. Das et al. \cite{LivDet2020} presented two different types of fake eyes: Van Dyke Eyes (which have higher iris quality details) and Scary Eyes (plastic fake eyes with a simple pattern on the iris region). They also presented add-ons for fake eyes, such as textured and clear contacts.

\begin{figure}[h]
  \centering
  \includegraphics[width=\linewidth]{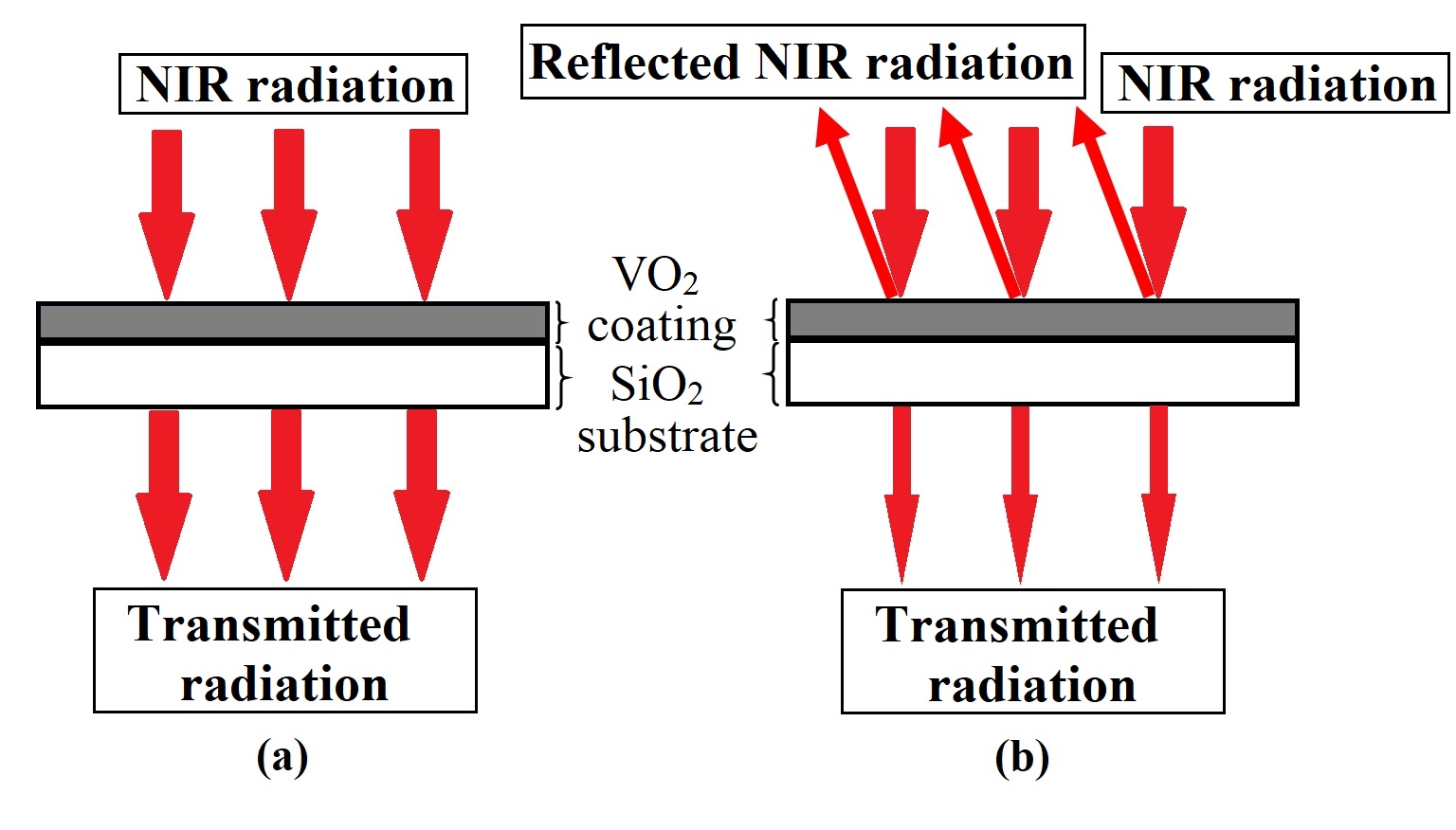}
  \caption{Schematic of thermochromic behavior of $\mathit{VO_2}$ films (a) below and (b) above 68\degree C (critical temperature).}
  \label{fig:vo2demo}
\end{figure}

As stated above, attacks using artificial eyes made of glass or plastic have not been heavily studied \cite{Chen2012, Lee2006, Zhang2011, Hoffman2018}.
The goal of this work is to leverage recent developments in material science to test the robustness and measure the susceptibility of iris systems to such type of spoof attacks. Specifically, we are interested to produce spoofs which are fabricated by affixing chemically modified films on these artificial eyes. 
 \footnote{In principle, it can be used on other types of PA artifacts.} The iris is generally imaged in the NIR spectrum; accordingly, we have attempted to use NIR-sensitive Vanadium dioxide ($\mathit{VO_2}$) films to generate these spoofs. $\mathit{VO_2}$ is a typical thermochromic material that has been widely studied as smart coatings for buildings fenestrations \cite{Zhang2011b, Gao2012b, Gao2012c, Gao2012a, Chang2018}. The synthesis of $\mathit{VO_2}$ films has been reported briefly in the literature, and its manufacturing is easy and cost effective. It is an advantage to use $\mathit{VO_2}$ for our work as it is deposited on a glass substrate and its handling is smooth. A further advantage is its low toxicity and high stability at room temperature conditions for such a short period of usage. These films show transmittance drop, close to a temperature of 68\degree C, in the NIR region \cite{Barker1966, Morin1959, Verleur1968, Lysenko2007a, Lysenko2007b}. This implies that at temperatures below 68\degree C, the film allows maximum light to pass through, but as the temperature increases above 68\degree C, the film behaves in a completely different manner, only allowing a portion of light to pass (Fig. \ref{fig:vo2demo}). This change in behavior of the film allows us to image the fake eyes in 2 different arrangements. Thus, in order to generate an effective spoof, we used the $\mathit{VO_2}$ coated and uncoated (blank) films in varied configurations on the fake glass eye. This is a unique kind of presentation attack, which combines multiple attack modes and that has never been attempted before.

\begin{figure}[h]
\centering
  \includegraphics[width=0.8\linewidth]{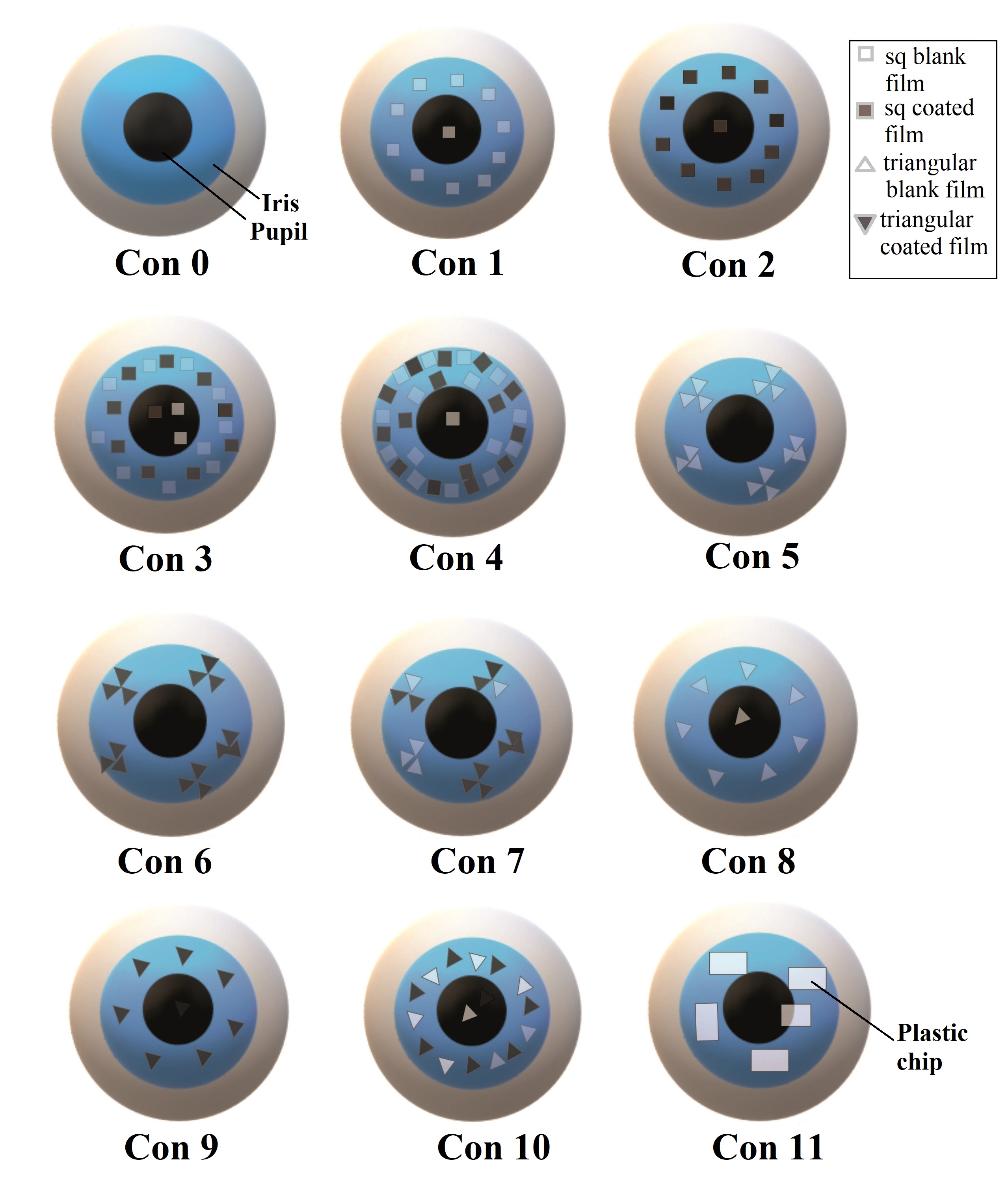}
  \caption{Diagrammatic representation of the various patterns in which the blank films and $VO_2$ coated films were arranged on the fake eyes.}
  \label{fig:Fig 1}
\end{figure}

\section{Experiment and Setup}

\subsection{Fabrication}
Vanadium dioxide ($\mathit{VO_2}$) thin films were deposited by pulsed laser deposition over fused silica ($\mathit{SiO_2}$) substrates (2" in diameter, 250 $\mu$m thick), following a process similar to what has been described in the past \cite{Coyetal-JAP-2010,Figueroaetal-AMT-2019}. The substrate was heated to 600$^ \circ$ C in a vaccum chamber with a background pressure of 1x10$^{-7}$ Torr. After reaching this set-point, oxygen and argon gas was introduced to the chamber, while the chamber pressure was controlled through a butterfly valve to be at 35 mTorr. At this point, laser pulses from an excimer laser (wavelength $\lambda$=248 nm, 20 ns pulse duration, and $\sim$ 4 J/cm$^{2}$ fluence) ablated a metallic vanadium target, and the pressure was controlled to 35 mTorr. A total of 320,000 pusles resulted in a 280 nm thick $\mathit{VO_2}$ thin film over the $\mathit{SiO_2}$ substrate. After $\mathit{VO_2}$ deposition, the 2" sample was diced into squares and triangles (2 mm $\times$ 2 mm). Another blank identical $SiO_{2}$ substrate (i.e., with no $\mathit{VO_2}$ thin film deposited) was also diced with the same dimensions. The resulting samples were multiple bare $\mathit{SiO_2}$ and $\mathit{VO_2}$-coated 2 mm $\times$ 2 mm ``pixels".  

\textbf{}
Our aim in this work is to fabricate fake eyes (Van Dyke eyes, made of soft glass) with different patterns of films on it. This patterning was based on different factors such as shape, type and orientation of the films (Fig. \ref{fig:Fig 1}). To achieve this, $\mathit{VO_2}$ coated films and blank films were fixed on the fake eyes in 11 different geometrical configurations as described below.   
For the first set of images (Con 0), the naked Van Dyke eyes were imaged in different angular and lighting conditions (Fig. \ref{fig:Fig 2} (a-j)). To achieve this, the Van Dyke eyes were first attached to fake Halloween glasses using double-sided tape. The user then mounted these glasses and approached the iCAM 7100S iris sensor for imaging. This triggered the activation of the sensor, as indicated by the appearance of an orange dot on the mirror. Now, at the correct distance, once the orange dot is aligned over the bridge of the nose, it turns green, and both the irides are acquired. This process was subsequently repeated by using the tilt up/down button on the sensor unit. Multiple other images (Fig. \ref{fig:Fig 2} i
(f-j)) were also captured by focusing some extra light (120 V, GE-IR table lamp) on the fake eyes (mounted on the user).   
For Con 1, a few blank square films were removed from the whole blank diced lot using a pair of tweezers. These films were then carefully stuck on the fake iris in a circular pattern (Fig. \ref{fig:Fig 2} ii (a-j)), with a couple of them on the pupil portion of the fake eyes. This patterned eye was imaged using the same process as stated above. One additional change was the {\em in situ} heating of the films using the IR lamp. The film was heated for 2.5 min to reach a temperature of 80\degree C, and a picture was acquired immediately. This was done to appreciate the difference in image and PA scores with and without heating (Fig. \ref{fig:Fig 2} ii (j)).    

A similar procedure was adopted for Con 2, where $\mathit{VO_2}$ coated square films were used instead of blank films (Fig. \ref{fig:Fig 2} (a-j)). Again, for Con 3, $\mathit{VO_2}$ coated and blank film were arranged alternately on the iris and pupil of the fake eyes (Fig. \ref{fig:Fig 2} iv (a-j)). Con 4 was designed by closely placing the coated and blank films in 2 rings on the iris, with one blank film on the pupil (Fig. \ref{fig:Fig 2} v (a-j)). Con 5 was fabricated by choosing triangular blank films. These triangular films were placed in a group of 3’s to form a flower-like pattern (Fig. \ref{fig:Fig 2} vi (a-j)). Similarly, $\mathit{VO_2}$ coated films were arranged in triangles of 3, forming flower-like pattern for Con 6 (Fig. \ref{fig:Fig 2} vii (a-j)). The Con 7 was designed using both coated and uncoated triangular films in grouping of 3 on the fake eyes (Fig. \ref{fig:Fig 2} viii (a-j)). Con 8, 9, and 10 were fabricated by placing triangular-shaped blank; triangular-shaped coated; and triangular-shaped blank and coated on the fake iris, respectively. (Fig. \ref{fig:Fig 2} ix-xi (a-j)). For the last configuration (Con 11), transparent plastic chips were stuck on the Van Dyke eyes (Fig. \ref{fig:Fig 2} (a-j)).
 
\begin{figure}[h]
  \includegraphics[width=\linewidth]{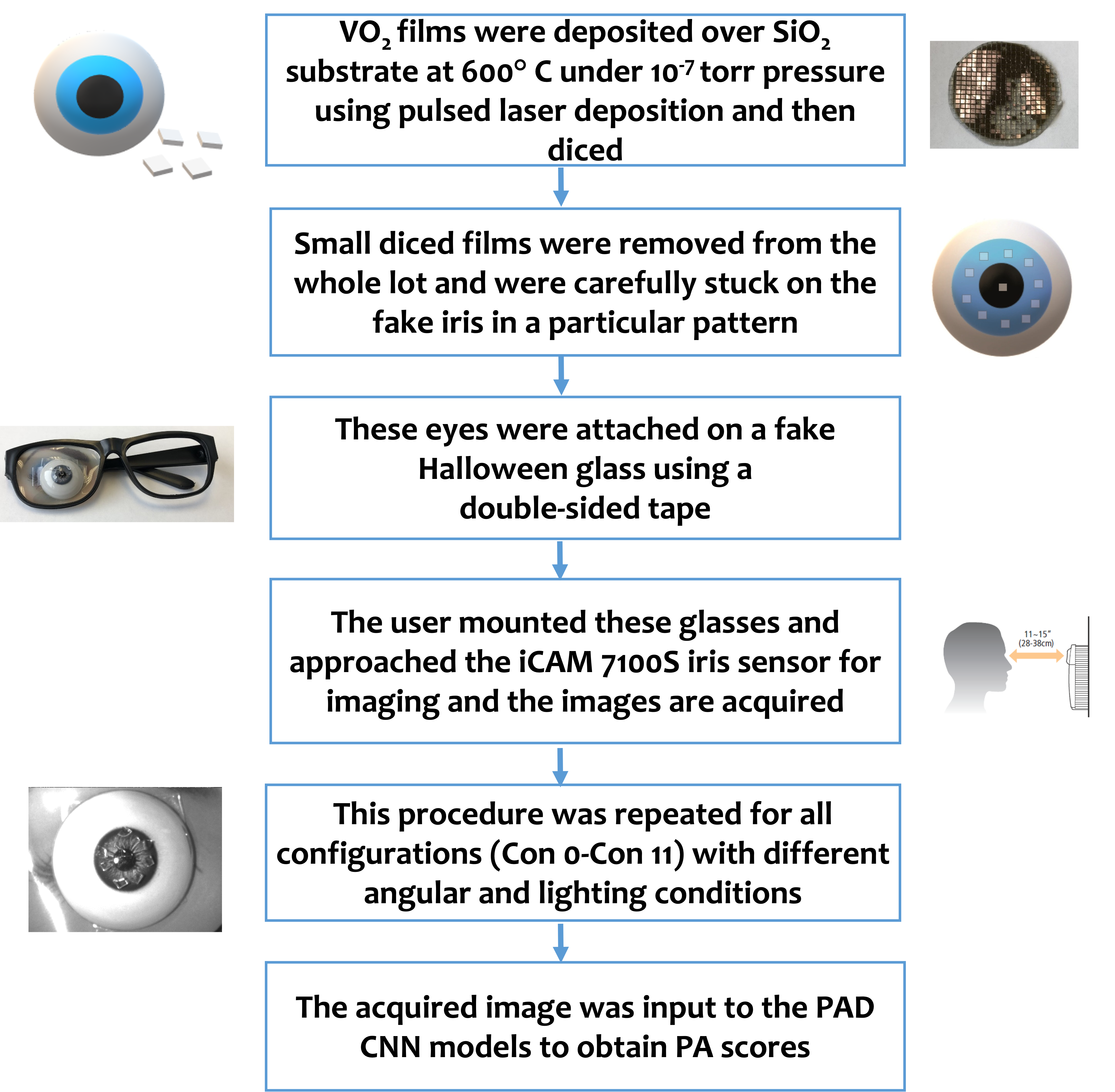}
  \caption{Step-wise procedure for fabrication of $\mathit{VO_2}$ modified fake eye starting from its deposition to PA score procurement \cite{iCAMRef}.}
  \label{fig:chart}
\end{figure}

\begin{figure*}[h]
  \includegraphics[width=\linewidth]{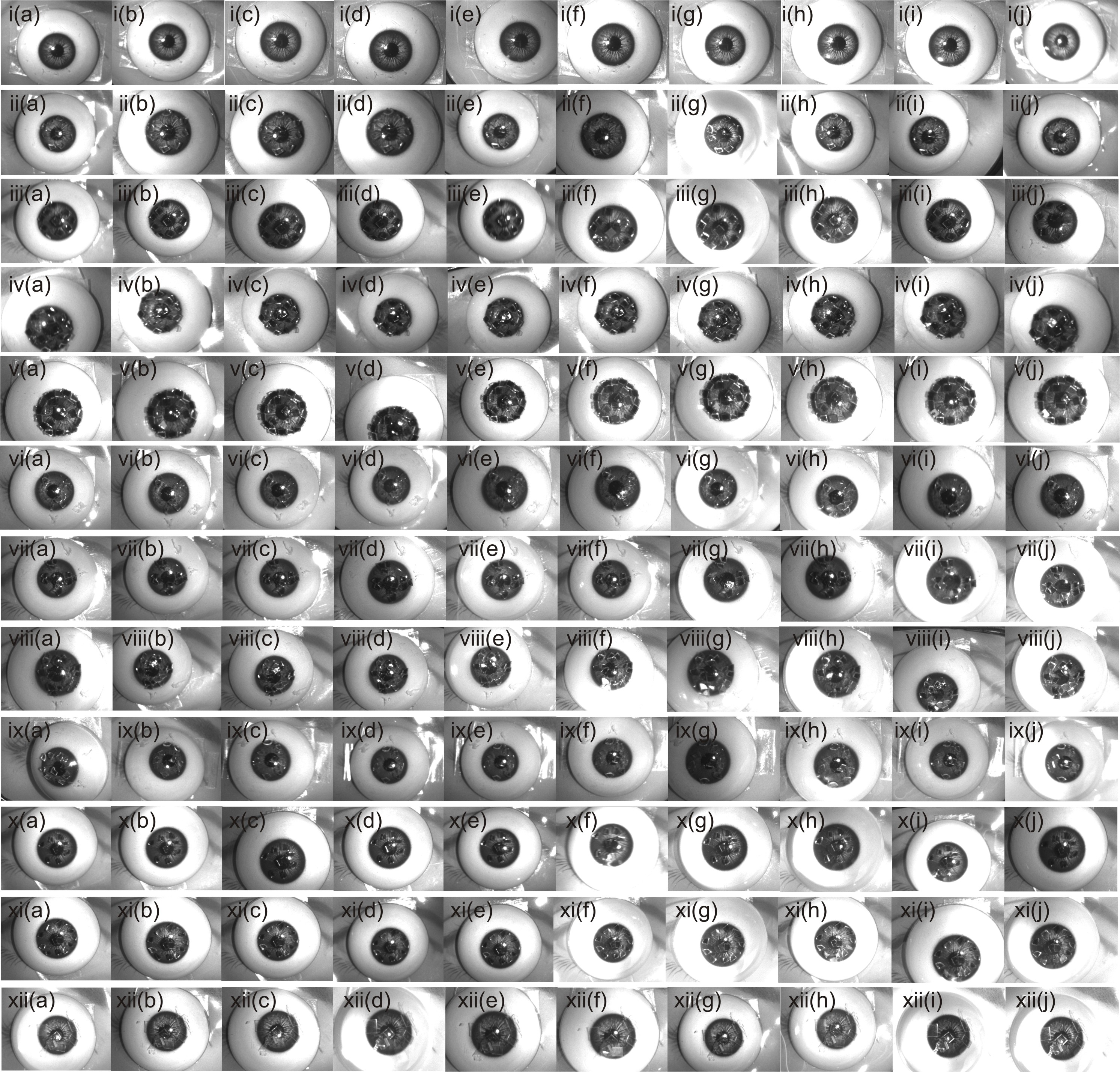}
  \caption{Images with various configurations of films as captured by iCAM 7100S. Images (a)-(j) represent a particular configuration taken in different angular, lighting and temperature conditions, sequentially (for a detailed label refer to Table 2). Images (i)-(xii) represent configurations Con 0 to Con 11.}
  \label{fig:Fig 2}
\end{figure*}

We captured 10 images of an eye for each configuration, resulting in a database of 120 samples. These images were taken with the help of six different subjects. More than one subject was used to eliminate any subject-specific errors during data collection. These images were then assessed using two state-of-the-art PA detectors: D-NetPAD and IrisTL-PAD (Fig. \ref{fig:chart}). Both PA detection methods produce a single-valued PA score. These PA scores range from 0 to 1, where 1 indicates a PA sample and 0 indicates a bonafide or live iris. 

\subsection{Iris Presentation Attack Detection Methods}

The two iris PA detection algorithms that are utilized to assess the vulnerability of adhering $\mathit{VO_2}$ films on artificial eyes are described below. They both are based on deep neural architectures.

\textbf{D-NetPAD}:
D-NetPAD \cite{Sharma2020}\footnote{https://github.com/iPRoBe-lab/D-NetPAD} is based on a densely connected convolutional neural network where each layer connects to every other layer in a feed-forward fashion. Its base architecture is DenseNet-121 \cite{Huang2017}, which consists of 121 convolutional layers in a series of four Dense Blocks and three Transition Layers. A detailed description of the architecture is provided in \cite{Huang2017}. To detect iris PA, the iris region is first cropped from the ocular image and resized to 224 $\times$ 224. The cropped and resized iris region is then input to the D-NetPAD, which produces a presentation attack (PA) score between 0 and 1. A flowchart of the D-NetPAD is shown in Fig. \ref{fig:D-NetPAD}. It utilizes a pre-trained ImageNet model to initialize weights and fine-tune them with iris PA samples. Fine-tuning has been performed with a proprietary dataset and the NDCLD2015 \cite{NDCLD2015} dataset. The proprietary dataset consists of 6,610 bonafide irides and 3,839 PA samples. PA samples include 130 Kindle replay attacks, 1,651 printed eyes, 1,537 plastic eyes, and 521 cosmetic contact lenses. From NDCLD2015 \cite{NDCLD2015}, 2,236 cosmetic contact lens images are used for training.

\begin{figure*}[h]
  \includegraphics[width=\linewidth]{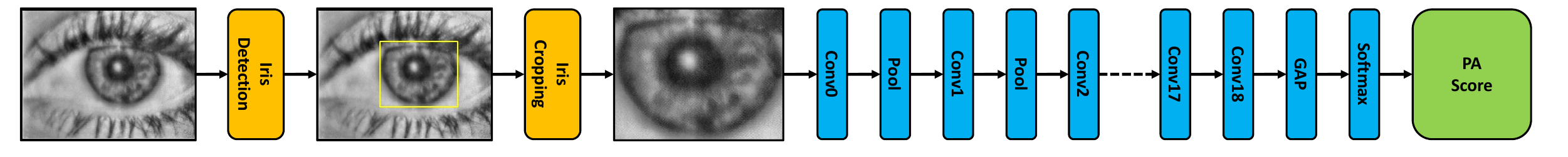}
  \caption{Flowchart depicting the IrisTL-PAD method.}
  \label{fig:boat1}
\end{figure*}

\begin{figure*}[h]
  \includegraphics[width=\linewidth]{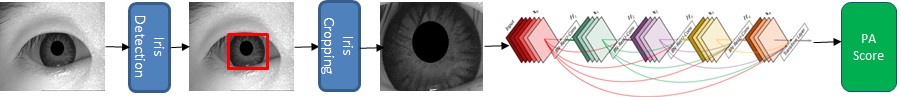}
  \caption{Flowchart of the D-NetPAD algorithm. Iris region (red box) is detected and cropped from the ocular image and input to the D-NetPAD architecture. The base architecture used in D-NetPAD is DenseNet-121 \cite{Huang2017}. It produces a single PA score which determines whether an input image is a bonafide or a PA.}
  \label{fig:D-NetPAD}
  \end{figure*}

\textbf{IrisTL-PAD}: IrisTL-PAD \cite{Chen2018a, ChenR18, Chen2021} operates on the cropped iris regions and offers a simple and fast solution for PA detection. It also utilizes the pre-trained ImageNet model to initialize the weights and then performs transfer learning. First, an off-line trained iris detector was used to obtain a rectangular region encompassing the outer boundary of the iris. Then, the iris region was automatically cropped based on the estimated rectangular coordinates. Finally, the cropped iris region was input to a CNN (ResNet50) to train the iris PA detection model (Fig. \ref{fig:boat1}). The training was fine-tuned on an existing ImageNet model, by leveraging extensive data augmentation schemes. The IrisTL-PAD model was trained on 9,072 bonafide images and 7,352 PA images as summarized in Table~\ref{datasets_tlpad}. 
\begin{table*}
\caption{A summary of datasets used to train IrisTL-PAD.}
\begin{center}
\begin{tabular}{l|c|c|c|c|c{c}r}
\hline
\textbf{Dataset} & \textbf{Total} & \textbf{Live} & \textbf{Print} & \textbf{Contact Lenses} & \textbf{Artificial Eye}   \\ \hline
LivDet-Iris 2017-IIT-WVU~\cite{LivDet2017} & 1,750 & 750 & - & 1000 & -   \\ \hline
LivDet-Iris 2017-NotreDame~\cite{LivDet2017} & 1,200 & 600 & - & 600 &  -   \\ \hline
LivDet-Iris 2017-Warsaw~\cite{LivDet2017} & 4,513 & 1,844 & 2,669 & - &  -   \\ \hline
BERC-Iris-Fake~\cite{BERC2007} & 4,598 & 2,778 & 1,600 & 140 &  80   \\ \hline
CASIA-Iris-Interval~\cite{HeLLLSH16} & 740 & - & - & 740 & - \\ \hline
Private Dataset & 3,623 & 3,100 & 6 & 334 & 183 \\ \hline
\textbf{Combined} &	\textbf{16,424} & \textbf{9,072} &	\multicolumn{3}{c}{\textbf{7,352}} \\ \hline
\end{tabular}
\end{center}
\label{datasets_tlpad}
\end{table*}


Both PAD algorithms are state-of-the-art methods that resulted in the best performance another proprietary dataset. The data were collected using the iCAM7000 NIR sensor from 1,315 subjects. A total of 3,315 iris images were acquired, out of which 2,963 were bonafide irides and 352 were PA samples. PAs in the dataset include two types of VanDyke eyes and 10 different types of cosmetic contact lenses. The D-NetPAD and IrisTL-PAD methods resulted in a True Detect Rate (TDR) of 98.58\% and 92.61\%, respectively, at a False Detect Rate (FDR) of 0.2\%.\footnote{ISO/IEC 30107-3:2023 specifies Attack Presentation Classification Error Rate (APCER) and Bonafide Presentation Classification Error Rate (BPCER) as evaluation metrics for PAD. TDR is 1$-$APCER, and FDR is the same as BPCER.} The TDR denotes the fraction of PA samples that were correctly classified, while the FDR denotes the fraction of bonafide samples that were incorrectly classified as PA samples. {\bf In addition, both PAD algorithms were the best performing algorithms in the LivDet-Iris 2020 competition \cite{LivDet2020}.}


\begin{table*}[]
\caption{Detailed table of PA scores for each image captured across all 12 configurations. Red colored cells represent PA scores for IrisTL-PAD which are less than or equal to its threshold value (0.5). Yellow colored cells represent PA scores for D-NetPAD which are less than or equal to its threshold value (0.4). Orange fonted numbers represent images taken with extra lightning conditions. Blue fonted numbers represent PA scores of images taken after heating the films.}
\label{table:Result08-12}
\resizebox{\textwidth}{!}{%
\begin{tabular}{|l|l|l|l|l|l|l|l|l|l|l|l|}
\hline
\multicolumn{2}{|l|}{} & \multicolumn{10}{c|}{PA scores} \\ \cline{3-12} 
\multicolumn{2}{|l|}{\multirow{-2}{*}{Configuration}} & a & b & c & d & e & f & g & h & i & j \\ \hline
 & IrisTL-PAD & 1.00 & 1.00 & 1.00 & 1.00 & 1.00 & {\color[HTML]{F8A102} 1.00} & {\color[HTML]{F8A102} 1.00} & {\color[HTML]{F8A102} 1.00} & {\color[HTML]{F8A102} 1.00} & {\color[HTML]{F8A102} 1.00} \\ \cline{2-12} 
\multirow{-2}{*}{Con 0 (Van Dyke eyes)} & D-NetPAD & 1.00 & 0.93 & 0.95 & 0.97 & 0.87 & {\color[HTML]{F8A102} 0.93} & {\color[HTML]{F8A102} 0.92} & {\color[HTML]{F8A102} 0.89} & {\color[HTML]{F8A102} 0.87} & {\color[HTML]{F8A102} 0.66} \\ \hline
 & IrisTL-PAD & 0.99 & 0.98 & 0.99 & 0.99 & 1.00 & {\color[HTML]{000000} 1.00} & {\color[HTML]{F8A102} 0.91} & {\color[HTML]{F8A102} 0.97} & {\color[HTML]{F8A102} 0.98} & {\color[HTML]{3531FF} \textbf{1.00}} \\ \cline{2-12} 
\multirow{-2}{*}{Con 1 (Blank sq all over)} & D-NetPAD & 0.66 & 0.57 & 0.58 & 0.60 & 0.57 & {\color[HTML]{000000} 0.64} & {\color[HTML]{F8A102} 0.57} & {\color[HTML]{F8A102} 0.59} & {\color[HTML]{F8A102} 0.56} & {\color[HTML]{3531FF} \textbf{0.65}} \\ \hline
 & IrisTL-PAD & 0.76 & 0.70 & 0.90 & 0.96 & 0.94 & 0.94 & {\color[HTML]{F8A102} 0.99} & {\color[HTML]{F8A102} 0.92} & {\color[HTML]{F8A102} 0.91} & {\color[HTML]{F8A102} 0.99} \\ \cline{2-12} 
\multirow{-2}{*}{Con 2 ($\mathit{VO_2}$ sq all over)} & D-NetPAD & 0.43 & 0.55 & 0.54 & 0.56 & 0.47 & 0.47 & {\color[HTML]{F8A102} 0.51} & {\color[HTML]{F8A102} 0.49} & {\color[HTML]{F8A102} 0.53} & {\color[HTML]{F8A102} 0.61} \\ \hline
 & IrisTL-PAD & \cellcolor[HTML]{FFCCC9}0.40 & \cellcolor[HTML]{FFCCC9}0.05 & \cellcolor[HTML]{FFCCC9}0.10 & \cellcolor[HTML]{FFCCC9}0.42 & \cellcolor[HTML]{FFCCC9}0.47 & \cellcolor[HTML]{FFCCC9}0.10 & \cellcolor[HTML]{FFCCC9}{\color[HTML]{F8A102} 0.01} & \cellcolor[HTML]{FFCCC9}{\color[HTML]{F8A102} 0.10} & \cellcolor[HTML]{FFCCC9}{\color[HTML]{3166FF} \textbf{0.23}} & \cellcolor[HTML]{FFCCC9}{\color[HTML]{3166FF} \textbf{0.09}} \\ \cline{2-12} 
\multirow{-2}{*}{Con 3 ($\mathit{VO_2}$ and Blank sq alternate)} & D-NetPAD & \cellcolor[HTML]{FFFFC7}0.35 & 0.48 & \cellcolor[HTML]{FFFFC7}0.37 & \cellcolor[HTML]{FFFFC7}0.38 & \cellcolor[HTML]{FFFFC7}0.35 & 0.43 & \cellcolor[HTML]{FFFFC7}{\color[HTML]{F8A102} 0.38} & \cellcolor[HTML]{FFFFC7}{\color[HTML]{F8A102} 0.39} & {\color[HTML]{3166FF} \textbf{0.41}} & {\color[HTML]{3166FF} \textbf{0.41}} \\ \hline
 & IrisTL-PAD & \cellcolor[HTML]{FFCCC9}0.16 & 0.92 & \cellcolor[HTML]{FFCCC9}{\color[HTML]{000000} 0.35} & {\color[HTML]{000000} 0.89} & 0.70 & {\color[HTML]{F8A102} 0.89} & \cellcolor[HTML]{FFCCC9}{\color[HTML]{F8A102} 0.25} & {\color[HTML]{F8A102} 0.73} & \cellcolor[HTML]{FFCCC9}{\color[HTML]{3166FF} \textbf{0.13}} & {\color[HTML]{3166FF} \textbf{0.60}} \\ \cline{2-12} 
\multirow{-2}{*}{Con 4 ($\mathit{VO_2}$ and Blank sq alternate ring)} & D-NetPAD & 0.42 & 0.43 & {\color[HTML]{000000} 0.43} & 0.43 & 0.42 & {\color[HTML]{F8A102} 0.42} & {\color[HTML]{F8A102} 0.44} & {\color[HTML]{F8A102} 0.43} & {\color[HTML]{3166FF} \textbf{0.43}} & \cellcolor[HTML]{FFFFC7}{\color[HTML]{3166FF} \textbf{0.40}} \\ \hline
 & IrisTL-PAD & \cellcolor[HTML]{FFCCC9}0.26 & 0.87 & 0.51 & \cellcolor[HTML]{FFCCC9}0.11 & \cellcolor[HTML]{FFCCC9}0.07 & {\color[HTML]{333333} 0.93} & {\color[HTML]{F8A102} 0.52} & \cellcolor[HTML]{FFCCC9}{\color[HTML]{F8A102} 0.11} & {\color[HTML]{3166FF} \textbf{0.95}} & \cellcolor[HTML]{FFCCC9}{\color[HTML]{3166FF} \textbf{0.13}} \\ \cline{2-12} 
\multirow{-2}{*}{Con 5 (Blank triangle in flower)} & D-NetPAD & 0.51 & 0.54 & 0.53 & 0.50 & 0.52 & {\color[HTML]{333333} 0.52} & {\color[HTML]{F8A102} 0.53} & {\color[HTML]{F8A102} 0.62} & {\color[HTML]{3166FF} \textbf{0.52}} & {\color[HTML]{3166FF} \textbf{0.46}} \\ \hline
 & IrisTL-PAD & 0.88 & 0.89 & \cellcolor[HTML]{FFCCC9}0.19 & \cellcolor[HTML]{FFCCC9}{\color[HTML]{000000} 0.02} & {\color[HTML]{F8A102} 0.80} & {\color[HTML]{F8A102} 0.99} & {\color[HTML]{F8A102} 0.55} & \cellcolor[HTML]{FFCCC9}{\color[HTML]{F8A102} 0.02} & {\color[HTML]{3166FF} \textbf{0.72}} & \cellcolor[HTML]{FFCCC9}{\color[HTML]{3166FF} \textbf{0.06}} \\ \cline{2-12} 
\multirow{-2}{*}{Con 6 ($\mathit{VO_2}$ triangle in flower)} & D-NetPAD & 0.53 & 0.47 & 0.49 & \cellcolor[HTML]{FFFFFF}{\color[HTML]{000000} 0.48} & {\color[HTML]{F8A102} 0.46} & \cellcolor[HTML]{FFFFC7}{\color[HTML]{F8A102} 0.40} & {\color[HTML]{F8A102} 0.44} & {\color[HTML]{F8A102} 0.46} & {\color[HTML]{3166FF} \textbf{0.54}} & {\color[HTML]{3166FF} \textbf{0.48}} \\ \hline
 & IrisTL-PAD & \cellcolor[HTML]{FFCCC9}{\color[HTML]{000000} 0.08} & \cellcolor[HTML]{FFCCC9}0.07 & \cellcolor[HTML]{FFCCC9}0.55 & \cellcolor[HTML]{FFCCC9}0.23 & \cellcolor[HTML]{FFCCC9}0.01 & \cellcolor[HTML]{FFCCC9}{\color[HTML]{F8A102} 0.03} & \cellcolor[HTML]{FFCCC9}{\color[HTML]{F8A102} 0.26} & \cellcolor[HTML]{FFCCC9}{\color[HTML]{F8A102} 0.44} & \cellcolor[HTML]{FFCCC9}{\color[HTML]{F8A102} 0.14} & \cellcolor[HTML]{FFCCC9}{\color[HTML]{3166FF} \textbf{0.23}} \\ \cline{2-12} 
\multirow{-2}{*}{Con 7 (Blank and $\mathit{VO_2}$ triangle in flower)} & D-NetPAD & \cellcolor[HTML]{FFFFC7}{\color[HTML]{000000} 0.31} & \cellcolor[HTML]{FFFFC7}0.35 & 0.43 & \cellcolor[HTML]{FFFFC7}0.37 & 0.47 & {\color[HTML]{F8A102} 0.44} & \cellcolor[HTML]{FFFFC7}{\color[HTML]{F8A102} 0.39} & {\color[HTML]{F8A102} 0.43} & \cellcolor[HTML]{FFFFC7}{\color[HTML]{F8A102} 0.38} & \cellcolor[HTML]{FFFFC7}{\color[HTML]{3166FF} \textbf{0.37}} \\ \hline
 & IrisTL-PAD & \cellcolor[HTML]{FFCCC9}0.24 & 0.92 & 0.90 & 0.70 & 0.73 & {\color[HTML]{F8A102} 1.00} & {\color[HTML]{F8A102} 0.60} & {\color[HTML]{F8A102} 0.99} & \cellcolor[HTML]{FFCCC9}{\color[HTML]{F8A102} 0.09} & \cellcolor[HTML]{FFCCC9}{\color[HTML]{F8A102} 0.01} \\ \cline{2-12} 
\multirow{-2}{*}{Con 8 (Blank triangle all over)} & D-NetPAD & 0.53 & 0.63 & \cellcolor[HTML]{FFFFC7}0.40 & 0.47 & 0.59 & {\color[HTML]{F8A102} 0.54} & {\color[HTML]{F8A102} 0.58} & {\color[HTML]{F8A102} 0.63} & {\color[HTML]{F8A102} 0.61} & {\color[HTML]{F8A102} 0.50} \\ \hline
 & IrisTL-PAD & 0.98 & 0.86 & \cellcolor[HTML]{FFCCC9}0.46 & \cellcolor[HTML]{FFCCC9}0.04 & 0.93 & {\color[HTML]{F8A102} 0.99} & {\color[HTML]{F8A102} 0.94} & {\color[HTML]{F8A102} 0.99} & {\color[HTML]{3166FF} \textbf{0.76}} & {\color[HTML]{3166FF} \textbf{0.70}} \\ \cline{2-12} 
\multirow{-2}{*}{Con 9 ($\mathit{VO_2}$ triangle all over)} & D-NetPAD & 0.67 & 0.68 & 0.64 & 0.61 & 0.60 & {\color[HTML]{F8A102} 0.75} & {\color[HTML]{F8A102} 0.62} & {\color[HTML]{F8A102} 0.61} & {\color[HTML]{3166FF} \textbf{0.65}} & {\color[HTML]{3166FF} \textbf{0.65}} \\ \hline
 & IrisTL-PAD & 0.59 & {\color[HTML]{000000} 0.89} & 0.79 & 0.97 & 1.00 & {\color[HTML]{F8A102} 0.96} & {\color[HTML]{F8A102} 0.94} & {\color[HTML]{F8A102} 0.93} & {\color[HTML]{F8A102} 0.67} & {\color[HTML]{3166FF} \textbf{0.83}} \\ \cline{2-12} 
\multirow{-2}{*}{Con 10 ($\mathit{VO_2}$ and Blank triangle all over)} & D-NetPAD & 0.48 & {\color[HTML]{000000} 0.44} & 0.44 & \cellcolor[HTML]{FFFFC7}0.40 & 0.42 & {\color[HTML]{F8A102} 0.50} & {\color[HTML]{F8A102} 0.44} & {\color[HTML]{F8A102} 0.57} & {\color[HTML]{F8A102} 0.47} & {\color[HTML]{3166FF} \textbf{0.47}} \\ \hline
 & IrisTL-PAD & 0.99 & 0.98 & 0.96 & 0.77 & 0.98 & {\color[HTML]{F8A102} 0.87} & {\color[HTML]{F8A102} 0.99} & {\color[HTML]{F8A102} 0.99} &{\color[HTML]{F8A102} 0.99} & {\color[HTML]{F8A102} 0.99} \\ \cline{2-12} 
\multirow{-2}{*}{Con 11 (Plastic chips sq all over)} & D-NetPAD & 0.50 & 0.51 & 0.58 & 0.51 & 0.53 & {\color[HTML]{F8A102} 0.51} & {\color[HTML]{F8A102} 0.49} & {\color[HTML]{F8A102} 0.44} & {\color[HTML]{F8A102} 0.55} & {\color[HTML]{F8A102} 0.53} \\ \hline
\end{tabular}}
\end{table*}

\section{Evaluation and Results }
\textbf{}
To determine whether the designed configurations of the Vanadium dioxide films on artificial eyes can be used to attack the system or not, we compared their PA scores to that of bonafide, i.e., live human eyes. The PA score for a live human eye ranges from 0.0-0.5 for IrisTL-PAD and 0.0-0.4 for D-NetPAD. As depicted in Table 2, Cons 0, 1 and 2 showed PA scores more than the threshold value (0.5 for IrisTL-PAD, 0.4 for D-NetPAD) for both the algorithms. This indicates that these configurations were detected as spoofs by both the algorithms. However, as we move onto Con 3, the PA scores dip below the threshold for all 10 images for IrisTL-PAD and 6 images for D-NetPAD. {\bf This is a successful configuration that fools the PA detection systems and passes as a live or bonafide eye} (Fig. \ref{fig:Fig 3}). Con 4, which has $\mathit{VO_2}$ coated and blank films in 2 concentric circles in the iris region, has an attack success rate of 40\% for IrisTL-PAD and 10\% for D-NetPAD (Fig. \ref{fig:Fig 3}). Attack success rate was calculated as the percentage of attacks below the given threshold value. A configuration having success rate of 50\% or more was considered to be a successful attack.  Con 5 which has triangular blank films arranged in a group of 3, shows attack success rate of 50\% for IrisTL-PAD and 0\% for D-NetPAD. Con 6 images have slightly lower chances of working as a spoof (rates: 40\% IrisTL-PAD and 10\% D-NetPAD). Con 7 on the other hand has higher chances of passing as a live eye, with attack success percentage at 90\% for IrisTL-PAD and 50\% for D-NetPAD. Con 8 has 30\% success for IrisTL-PAD, and 10\% for D-NetPAD. Con 9 too has a lower chance to deceive the system (only 20\% success for IrisTL-PAD and none for D-NetPAD). Cons 10 (10\% success rate for D-NetPAD) and 11 also do not pose a threat to the two PA detection methods. One point to be mentioned is that the heating of $\mathit{VO_2}$ films upto a temperature of 80\degree C does not bring a significant change in the PA scores. Cons 1(j), 5(i), 9(i), 9(j) and 10(j) show high PA scores even when the films are above the critical temperature, and reflecting most of the light. This is an indication that the thermochromic behaviour of the film does not play a big role in deceiving the system as far as our experimental protocol is concerned.

\begin{figure}[h]
  \includegraphics[width=\linewidth]{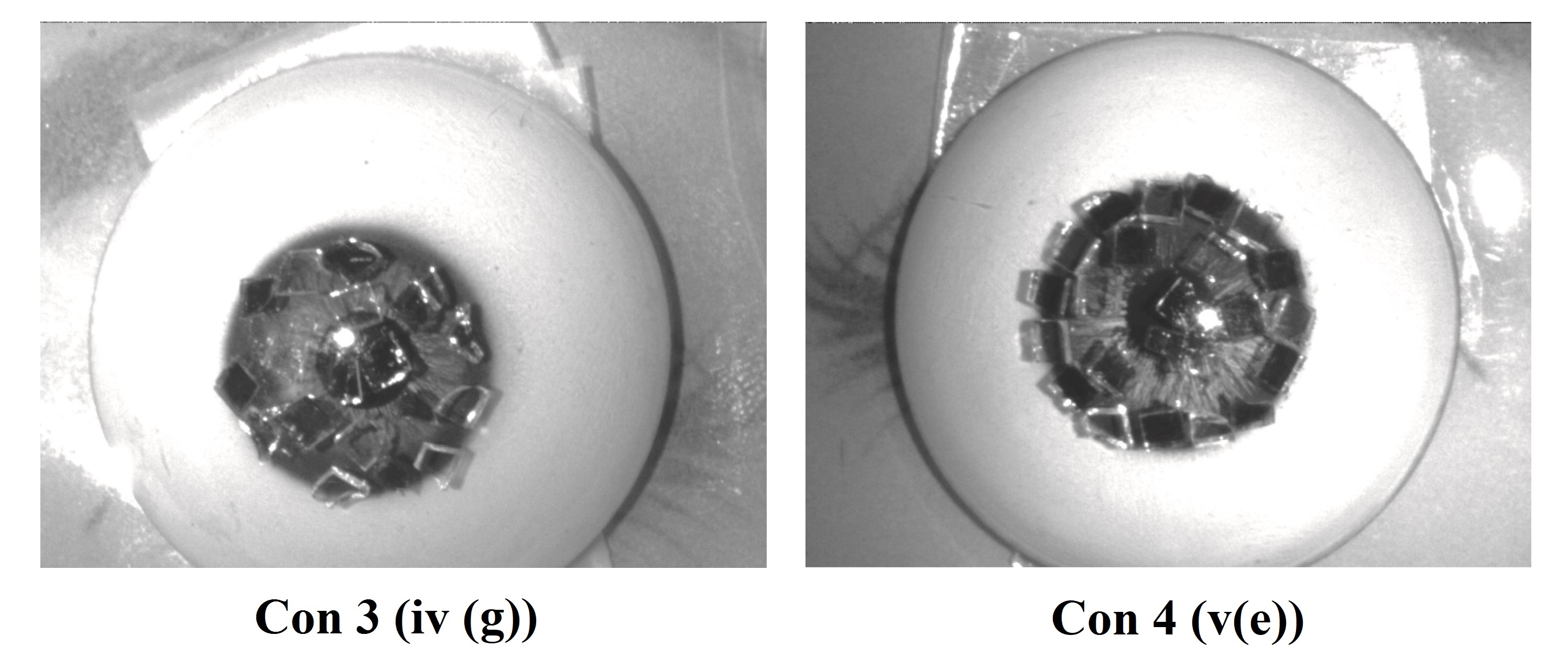}
  \caption{Comparison of the geometrical arrangement of Cons 3 and 4. Con 3 has $\mathit{VO_2}$ coated and blank films arranged in an alternate manner, but in no particular geometrical pattern all over the artificial eye. Con 4 on the other hand, has these films arranged in 2 concentric circles inside the iris region and 2 blank films on the pupil of the fake eyes.}
  \label{fig:Concomp}
\end{figure}

\begin{figure}[h]
\centering
  \includegraphics[width=0.8\linewidth]{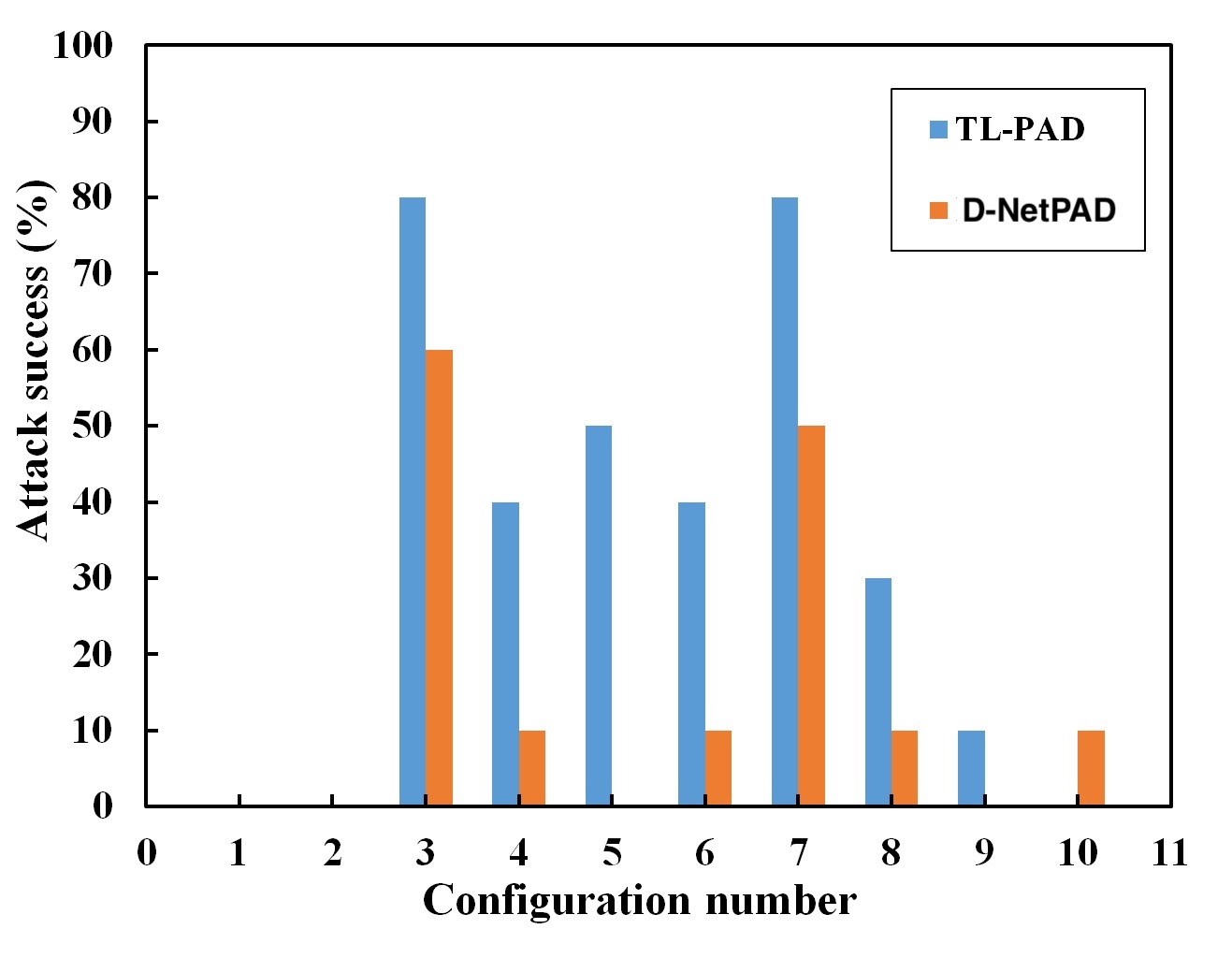}
  \caption{Presentation attack success rate across all configurations. Cons 3, 5 and 7 have a higher success rate (against the IrisTL-PAD method) compared to other configurations. This suggests that the films may have to be strategically placed on the fake iris pattern in order to defeat an iris PA detection system.}
  \label{fig:Fig 3}
\end{figure}

{\bf In summary, our preliminary observations indicate that Cons 3, 5 and 7 have high presentation attack success rates}. The high presentation attack success rate for Cons 3, 5 and 7 could be due to the kind of geometrical arrangement of films on them. Note that Cons 3 and 4 have a similar type of arrangement for both the films (coated and uncoated), but Con 3 has more space between the films (Fig. \ref{fig:Concomp}). This causes a change in the captured iris pattern and impacts the PA detection methods.    

\begin{figure}[h]
  \includegraphics[width=\linewidth]{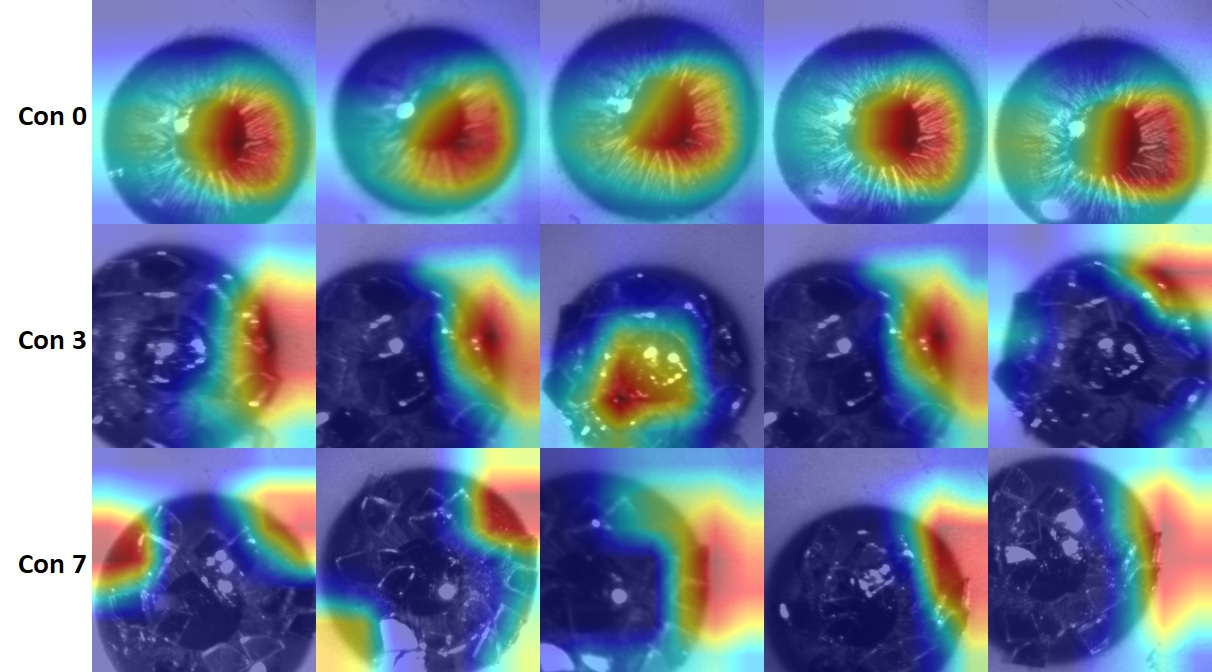}
  \caption{Grad-CAM \cite{Selvaraju2017} heatmaps of images corresponding to Con 0, Con 3 and Con 7 configurations. Con 0 has low PA success rate, whereas Con 3 and Con 7 have a high PA success rate. Red-colored regions represent highly focused region by the D-NetPAD. The blue region represents low priority regions. These regions help in making the final decision about being a bonafide or a PA.}
  \label{fig:Grad-CAM}
\end{figure} 

The result was further visually analyzed by generating ``heatmaps" using Gradient-weighted Class Activation Mapping (Grad-CAM) \cite{Selvaraju2017}. Grad-CAM produces a coarse localization map highlighting the salient regions in an image that were used by the network in order to generate its inference. Fig. \ref{fig:Grad-CAM} presents the ``heatmaps" for configurations that were unsuccessful (Con 0) as well as those that were successful (Con 3 and Con 7) in defeating the D-NetPAD algorithm. The red regions indicate high activation, whereas the blue regions represent low activation when inferring the final decision (i.e., bonafide or PA). The first row of Fig. \ref{fig:Grad-CAM} shows the heatmaps of Con 0 images, where the high activation region is at the pupillary zone of the printed iris pattern of the fake eye. The other two rows of Fig. \ref{fig:Grad-CAM} correspond to Cons 3 and 7 (high presentation attack success rate). The high activation regions in these two rows of images are distributed throughout the iris pattern.

We hypothesize that the combination of $\mathit{VO_2}$ and blank films when placed on the Van Dyke eyes interfered with the iris pattern inscribed on the fake eye. This presumably resulted in a pattern that was never seen by the algorithm during training. As a result, the focus shifted away from the iris pattern (see the last two rows of Fig. \ref{fig:Grad-CAM}), resulting in PA scores that were in the vicinity of the threshold (0.40). The chances of mis-classification seems to have increased with an increase in the density of the $\mathit{VO_2}$ and blank films. The $\mathit{VO_2}$ films appear to obscure the underlying pattern due to their special optical property with NIR illumination, whereas the blank films distort the pattern. Thus, Cons 3 and 7, which have a high concentration of $\mathit{VO_2}$ and blank films (Fig. \ref{fig:Fig 2}), show high misclassification rates.

\begin{figure}[h]
  \includegraphics[width=\linewidth]{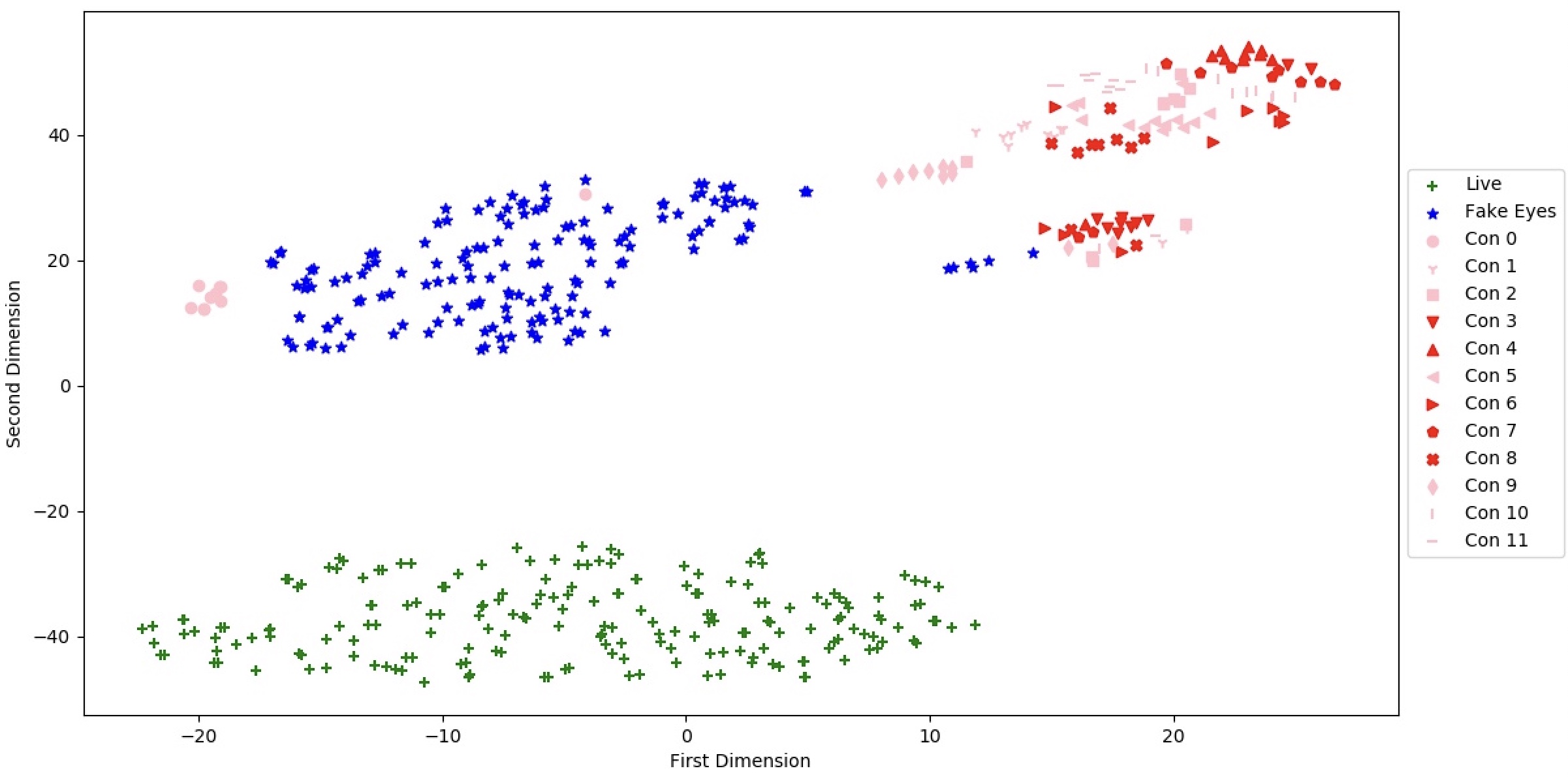}
  \caption{t-SNE plot of the D-NetPAD algorithm on bonafide (green) and fake eye (blue) images on the proprietary dataset. It also shows the t-SNE of all the configurations considered in this work. Red-colored data points represent configurations with a high PA attack success rate (Con 3, 4, 6, 7, 8), whereas pink represents configurations with a low PA attack success rate (Con 0, 1, 2, 5, 9, 10, 11). The distribution of the configurations is observed to be substantially different from that of the fake eyes, suggesting the novel nature of the attack.}
  \label{fig:D-NetPAD-tsne}
\end{figure}

After Grad-CAM visualization, which utilizes back-propagation, we also visualized the features fed into the architecture in the forward direction for the final decision. The features were extracted from the penultimate layer (just before the fully connected layer) of D-NetPAD and reduced to two dimensions using t-Distributed Stochastic Neighbor Embedding (t-SNE) \cite{Maaten2008}. t-SNE plots are shown in Fig. \ref{fig:D-NetPAD-tsne}, where the green and blue data points represent bonafide and fake eye images from the proprietary dataset, respectively. The red data points (Fig. \ref{fig:D-NetPAD-tsne}) represent configurations with a high PA success rate (Cons 3, 4, 6, 7, 8), whereas the  pink data points represent configurations with a low PA success rate (Cons 0, 1, 2, 5, 9, 10, 11). Fig. \ref{fig:D-NetPAD-tsne} shows the distribution of the configurations departing from that of the fake eyes, as well as being spaced out. This divergence further substantiates the effectiveness of using $\mathit{VO_2}$ films in performing iris presentation attacks.

\section{Role of Coated and Blank Films}
It is clear from the experiments conducted in this work that introduction of the $\mathit{VO_2}$ films along with blank films made a difference in the optical properties of the fake eyes. This change triggered a shift of focus of the PAD algorithms away from the iris portion, labelling them as bonafide. To check the function of $\mathit{VO_2}$ films, we carried out some preliminary experiments using metal coated films. These films when used alternately with blank films on fake eyes lowered the PA scores. This clearly shows that the new films (metal), just like $\mathit{VO_2}$ ones, caused changes in the optical properties of the fake eyes. These films worked as an attack only when used with blank films but not just by themselves. This suggests that a patch-based configuration is able to fool the PAD algorithm and pass as a genuine eye. But extensive experiments have to be carried out with the new films which can help us strengthen this hypothesis.   

\section{Mitigation Measures}
We performed another experiment to find a potential solution for the misclassification of $\mathit{VO_2}$ and blank films coated fake eyes as bonafides. We utilized all samples (=10) from a subset of configurations defined in this paper to perform incremental training of the D-NetPAD algorithm. The configurations used for the training were 1, 2, 5, 9 and 11 as they were correctly classified as PAs by the D-NetPAD. For incremental training, we fine-tuned the D-NetPAD model with the selected samples. Next, we recomputed the PA scores of all samples, including those pertaining to Cons 3, 4, 6, 7, 8, 10 which were not used for training. We observed that all the samples were now correctly classified as PAs. The experiment shows that incremental training with only a few samples can extend the discriminative power of the model in detecting such new attacks.

\section{Discussion and Future Work}
In this work, we assessed the possibility of combining Vanadium dioxide films with artificial glass eyes in order to create PAs that can potentially evade presentation attack detection. $\mathit{VO_2}$ films can be used to selectively regulate NIR transmission thereby causing such artificial eyes to be misclassified as bonafide samples. Our experimental results suggest that the placement of these films in specific configurations can indeed confound a PA detection system.

Having said that, there are some ways to detect these types of attacks: (a) Patch-based PAD: The Vanadium dioxide films used to create the PAs, modified the iris texture in configurations that can be described by local patches (see Fig. \ref{fig:Fig 2}). Both IrisTL-PAD and D-NetPAD solutions extract global features from the cropped iris images. By using local regions for PA determination, it is likely that patches which are not modified with $\mathit{VO_2}$ films would produce high PA scores. Hence, averaging the PA scores across individual patches can increase the robustness of these PAD solutions to the proposed attack \cite{Hoffman2019}. 
(b) One-class Classification: It is difficult to model the distribution of every unknown or unseen PAs. To tackle such PAs, a one-class classifier concept can be leveraged where only bonafide distribution is required to create the PAD model \cite{Yadav2020}.

Future work will explore the application of Vanadium dioxide films to generate PA artifacts that can be used for training and increase the robustness of existing PAD methods. Further, their thermochromic behavior can possibly be used to design new iris hardware for PA detection. We will also study the efficacy of this attack on other PAD techniques. In this work, the attack has been studied in the context of PAD only; future work will involve analyzing the impact on the iris recognition method also.

\textit{\textbf{Ethical Implications}: The goal of this paper was to alert researchers and practitioners to potential attacks and provide a preliminary solution to detect them. However, it should not be misused to launch an attack against iris recognition systems.}




{\small
\bibliographystyle{ieee_fullname}
\bibliography{main}
}

\balance

\end{document}